\documentclass[twocolumn,10pt,journal,twoside]{IEEEtran}
\usepackage[cmex10]{amsmath}
\usepackage{amssymb}
\usepackage{amsfonts}
\usepackage{amsthm}
\usepackage{algorithm}
\usepackage{algpseudocode}
\usepackage{graphicx}
\usepackage{epsfig}
\usepackage{cite}
\usepackage{tensor}
\usepackage[caption=false,font=footnotesize]{subfig}
\usepackage{color}
\usepackage{makecell}
\usepackage{float}
\usepackage{comment}
\usepackage{subfig}
\usepackage[dvipsnames]{xcolor}

\usepackage{hyperref}
\hypersetup{hypertex=true,
	colorlinks=true,
	linkcolor=blue,
	anchorcolor=blue,
	citecolor=blue}

\graphicspath{{figures/}}
\DeclareGraphicsExtensions{.eps}
\theoremstyle{definition}
\newtheorem{remark}{Remark}

\newcommand\T{{\hspace{-0pt}\intercal}}

\begin{document}

\title{
Bezier-based Regression Feature Descriptor for Deformable Linear Objects
}

\author{
Fangqing Chen $^*$\\
University of Toronto

\thanks{
Copyright may be transferred without notice, after which this version
may no longer be accessible.
}
\thanks{
$^*$ Corresponding Author.
}

}

\bstctlcite{IEEEexample:BSTcontrol}

\maketitle

\begin{abstract}
In this paper, a feature extraction approach for the deformable linear object is presented, which uses a Bezier curve to represent the original geometric shape.
The proposed extraction strategy is combined with a parameterization technique, the goal is to compute the regression features from the visual-feedback RGB image, and finally obtain the efficient shape feature in the low-dimensional latent space.
Existing works of literature often fail to capture the complex characteristics in a unified framework.
They also struggle in scenarios where only local shape descriptors are used to guide the robot to complete the manipulation.
To address these challenges, we propose a feature extraction technique using a parameterization approach to generate the regression features, which leverages the power of the Bezier curve and linear regression.
The proposed extraction method effectively captures topological features and node characteristics, making it well-suited for the deformation object manipulation task.
Large mount of simulations are conducted to evaluate the presented method.
Our results demonstrate that the proposed method outperforms existing methods in terms of prediction accuracy, robustness, and computational efficiency.
Furthermore, our approach enables the extraction of meaningful insights from the predicted links, thereby contributing to a better understanding of the shape of the deformable linear objects.
Overall, this work represents a significant step forward in the use
of Bezier curve for shape representation.
\end{abstract}


\IEEEpeerreviewmaketitle

\section{Introduction}\label{section1}
The manipulation of deformable objects is currently an open (and hot) research problem in robotics \cite{zaidi2015interaction} that has attracted many researchers due to its great applicability in many areas, e.g. manipulating fabrics \cite{torgerson1988vision}, shaping of food materials \cite{van2022incorporating}, assembling soft components \cite{kennaway2011generation}, manipulating cables \cite{halt2018intuitive}, interacting with tissues \cite{chen2021vx}, etc.
Note that physical interactions between a robot and a deformable object will inevitably alter the object's shape.
The feedback control of these additional object degrees-of-freedom (DOF) is referred to in the literature as shape servoing \cite{yang2023integrating}, a frontier problem that presents one main challenge \cite{qi2022towards}: 
The efficient feedback characterization of the object's shape \cite{Hou2019A,Foresti2004Automatic}, which has an infinite number of DOF to be controlled by a robot with limited manipulation directions \cite{ma2022active,2010Dexterous}.

Bezier curve, a class of fitting models designed to work with shape fitting and approximation, has demonstrated remarkable success in numerous tasks such as shape fitting \cite{pastva1998bezier,gousenbourger2019data}, data extraction \cite{d2018multi,bansal2015emotion}, and shape reconstruction \cite{galvez2008particle,nugroho2015reconstruct}.
Their ability to capture both local and global structural information, as well as node features, makes them a promising candidate for the problem of shape representation in the field of deformable object manipulation \cite{perez2020speed,li2006reconstruction}.

In this paper, we introduce a Bezier-based shape representation framework that leverages the power of regression features and linear regression,
Our approach extends traditional Bezier curve models by incorporating parameterization curve enhancements tailored specifically to the unique challenges presented by shape representation.

To the best of our knowledge, this is the first attempt to design a feature extraction framework for DLO with the consideration of the Bezier curve and parameterization technique, which helps to obtain a low-dimensional shape descriptor in the latent feature space.

The rest of this paper is organized as follows: 
Section 2 provides a review of related work in the field of shape representation and Bezier curve. 
Section 3 presents the methodology of the regression feature extraction, including the problem formulation and the algorithm architecture. 
Section 4 describes the simulation setup and presents the results and comparison to existing methods. 
Finally, Section 6 concludes the paper and discusses potential directions for future work.

The key contributions of this paper are two-fold:
\begin{itemize}
\item 
\textbf{Bezier-based Shape Descriptor.}
This paper forms a Bezier-based extraction framework utilizing the parameterization regression.
The core modules are constructed in three parts, the build of the cost function, the build of the augmented function, and the linear regression,
The module aims to generate a low-dimensional feature representing the original shape configuration of the deformable linear object.
The proposed manipulation runs in a model-free manner and does not need any prior knowledge of the system model.

\item 
\textbf{The Linear Regression Calculation.}
This paper uses linear regression to solve the fitting cost function, thereby directly obtaining the analytical expression of the shape feature and also providing the conditions for the effective stability of the feature.
Moreover, the calculation method has good real-time performance and stronger robustness than the other counterparts.
\end{itemize}

\section{Related Work}
In this section, we review the existing literature relevant to our study.
This discussion is divided into two main subsections:
shape detection and feature extraction.

\subsection{Shape Detection}
In the field of deformable object manipulation, the important module is to obtain a suitable shape description \cite{hayen2018high,vranic2002description,ronse1991morphological}.
For this issue, the current methods are mainly divided into two aspects:
local shape descriptor (LSD) and global shape descriptor (GSD).
As for LSD, the typical example is the point form in 2D/3D points, e.g., central point, marker point, feature point, etc.
This point form is the simplest LSD using data points as shape features of deformable linear objects. 
Because no matter from the perspective of geometry or image processing, features based on data points are the simplest features that can be extracted.
For example, feature points \cite{shademan2004using,mcfadyen2016image}, 
hole points \cite{wang2008micropeg,lu2023cfvs,liu2020hybrid}, and other artificial markers \cite{vicente2017towards,luo2020natural}.
However, point-based LSD has an obvious drawback, i.e., high-dimensional considerable, and low robustness to lightness, contrasts, and other human interference \cite{ardon2018reaching}, which can easily lead to inaccurate DLO shape description and affect system control accuracy \cite{peng2020comparing,zeng2023adaptive}.
To this end, the center-mass point is used to represent the shape change of the ball's middle area in \cite{nuriyev2007simulation}, which improves the anti-interference of the point-based feature.
Other types of LSD,  such as lines, ellipses, etc., and hybrid features are also used in some servo tasks \cite{zou2022deep}.
However, the above-point-based features often face the high-dimensional question, which may have a bad effect on the real-time system's performance, and even cause the manipulation failure \cite{marey2008analysis}.
With the above talk, it is crucial to design a feature extraction method for the original shape configuration to generate a low-dimensional feature vector, which captures the physical characteristics as much as possible \cite{zeng2022adaptive}.
Especially in some specified cases with highly dynamic processes, the shapes change fast, and an efficient shape descriptor is needed in complex environments.
Meanwhile, it is also important to enhance the robustness to the problems of illumination, contrast, environmental noise, and object surface texture in the real environment.

\subsection{Feature Extraction}
For the traditional rigid objects, it is easy to use 6-DOF vectors to describe the whole shape configuration \cite{laporte2006rigid}.
However, as flexible objects usually have infinite-dimensional geometric information, thus it is not possible to use simple 6-DOF to represent \cite{samangooei2008use}.
Thus, a simple and effective feature extractor that can characterize these objects in an efficient (i.e., compact) manner should be designed \cite{tawbe2016data}.
A method based on linearly parameterized (truncated) Fourier series was also proposed to represent the object's contour \cite{stromberg1986fourier}.

\section{Beizer-based Feature}
In this article, the centerline configuration of the object is defined as follows:
\begin{align}
\label{eq1}
\bar{\mathbf{c}}=
[\mathbf{c}_1^\T,\ldots,\mathbf{c}_N^\T]^\T \in \mathbb{R}^{2N}, \ \ \mathbf{c}_i=[c_{xi},c_{yi}]^\T \in \mathbb{R}^2
\end{align}
where $N$ is the number of points comprising the centerline, $\mathbf{c}_i$ for $i=1,\ldots,N$ are the pixel coordinates of $i$-th point represented in the camera frame.

Note that the dimension $2N$ of the observed centerline $\bar{\mathbf{c}}$ is generally large, thus it is inefficient to directly use it in a shape controller as it contains redundant information.
In this work, we just give this generation concept, i.e., shape feature extraction.
This module aims to extract efficient features $\mathbf{s}$ from the original data space and then map them into the low-dimensional feature space.

In this work, we regard the centerline $\bar{\mathbf{c}}$ of the 2D feedback as a continuous parametric curve of arc length with the following format:
\begin{equation}
\label{eq2}
\mathbf{c}_i = \mathbf{f}(\rho_i) \in \mathbb{R}^2 , \ \ \ i = 1, \ldots, N
\end{equation}
where $\rho \in [0,1]$ is a parametric variable representing the curve's normalized arc length.
In the level of the geometric meaning, ${\rho}_i$ can be seen as the arc-length between the start point $\mathbf{c}_1$ and point $\mathbf{c}_i$, e.g., $\rho_1=0$ and $\rho_N=1$.
The Bezier-based parametric curve fitting is modeled as follows:
\begin{equation}
\label{eq3}
{\mathbf{f} \left( {{\rho_i}} \right) = \sum\limits_{j = 0}^n {{\mathbf{p} _j}{B_{j,n}}\left( {{\rho_i}} \right)}}, \ \ \
i = 1, \ldots, N
\end{equation}
where $n\in\mathbb N$ is the fitting order, and $\mathbf{p}_j \in \mathbb{R}^2$ represents the shape parameters of $\bar{\mathbf{c}}$, and $B_{j,n}(\rho)$ is the chosen Bezier regression parameterization, which can take the following form:
\begin{align}
\label{eq4}
B_{j,n}\left( \rho \right) = \frac{n!}{j!\left( n - j \right)!} \left(1 - \rho  \right)^{n - j}\rho ^j
\end{align}
Bezier curve approximates the curve with a polynomial expression using control points. 
$n+1$ control points can determine a $n$-degree Bezier curve.
Interestingly, the feature vector $\mathbf{p}$ generated from the Bezier curve has the obvious physical meaning, i.e., the shape control points of the Bezier curve.
Bezier curve has a first-order derivability, and it can guarantee that the fitting will advance smoothly with the control points without fluctuations, thus, it can represent complex shapes.
However, there may be a computational burden as the fitting order increases.

By using the parameterization curve \eqref{eq2} and Beizer fitting mode \eqref{eq3}, the cost function can be constructed as follows:
\begin{equation}
\label{eq5}
Q = \sum\limits_{i = 1}^N {{{\| {{\mathbf{c}_i} - \sum\limits_{j = 0}^n {{  \mathbf{p}_j}{B_{j,n}}\left( {{\rho _i}} \right)} } \big\|^2}}} 
\end{equation}

Then, the augmented matrix format, refer to \eqref{eq5} can be further improved as follows:
\begin{equation}
\label{eq6}
Q = {\left( {\mathbf{B}\mathbf{s} - \bar{\mathbf{c}}} \right)^\T}
\left( {\mathbf{B}\mathbf{s} - \bar{\mathbf{c}}} \right)
\end{equation}
for a ``tall'' regression matrix $\mathbf B$ satisfying:
\begin{align}
\label{eq7}
\mathbf{B} &= [\mathbf{B}_1^\T, \ldots, \mathbf{B}_N^\T]^\T 
\in \mathbb{R}^{2N \times 2(n+1)} \notag \\
\mathbf{B}_i &= [B_{0,n}(\rho_i),\ldots,B_{n,n}(\rho_i)] \otimes \mathbf{I}_2 \in \mathbb{R}^{2 \times 2(n+1)}  
\end{align}
where $\mathbf{I}_2$ is the eye matrix with format $2 \times 2$, and $\otimes$ is the Kronecker operation.

We seek to minimize \eqref{eq6} to obtain a feature vector $\mathbf{s}$ that closely approximates
Thus, the feature vector $\mathbf{s}$ is computed using normal equation at every iteration as:
\begin{equation}
\label{eq68}
\mathbf{s} = {\left( {{\mathbf{B}^\T}\mathbf{B}} \right)^{ - 1}}{\mathbf{B}^\T}\bar{\mathbf{c}}
\end{equation}
where it is assumed that $N\gg n+1$ is should be met to ensure the reversibility of $\mathbf{B}^{\T} \mathbf{B}$.

\begin{remark}
\label{remark1}
Although this paper only gives the Bezier curve form, many other geometric expressions can be used as the regression, e.g., B-spline and rational approximation.
\end{remark}

\begin{figure}[ht]
\centering
\subfloat[GMM]
{\includegraphics[scale=0.71]{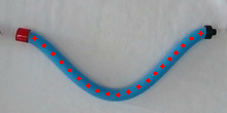}}
\subfloat[FCM]
{\includegraphics[scale=0.71]{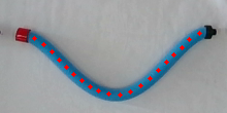}}
	
\subfloat[KMS]
{\includegraphics[scale=0.71]{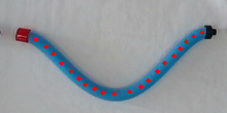}}
\subfloat[SOM]
{\includegraphics[scale=0.71]{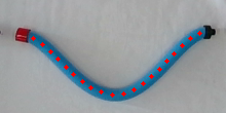}}
	
\caption{
Comparison of shape extraction effects of four clustering algorithms
within centerline configuration.
}
\label{fig1}
\end{figure}

\section{Experimental Results}
In this section, several experiments are conducted to validate the effectiveness of the proposed feature extraction method.

\subsection{Centerline Detection}
For accurate shape detection, we adopt the method in [x] to give examples.
Four clustering algorithms are used to detect the centerline configuration from the original 2D feedback RGB image.
The adopted methods are
Self Organization Map (SOM),
K-means (KMS),
Gaussian Mixture Model (GMM)
Fuzzy C-Means (FCM).

From Fig. \ref{fig1}, it can be seen that KMS has the best clustering ability to extract the centerline.
Thus, throughout the paper, the KMS technique is used as the centerline extraction measure.

\begin{figure}[ht]
\centering
\subfloat[GMM]
{\includegraphics[scale=0.19]{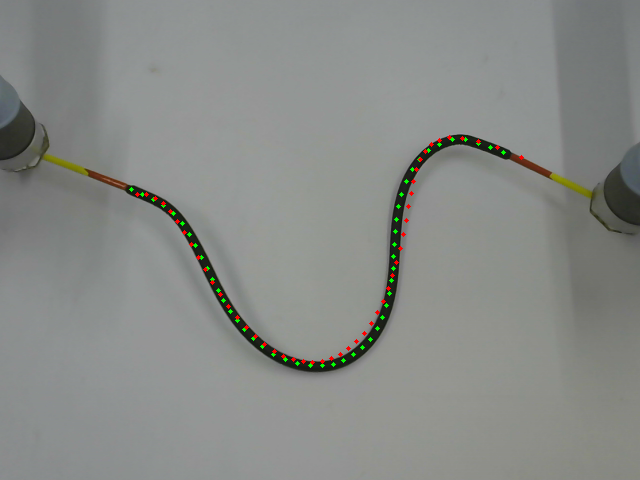}}
\subfloat[FCM]
{\includegraphics[scale=0.19]{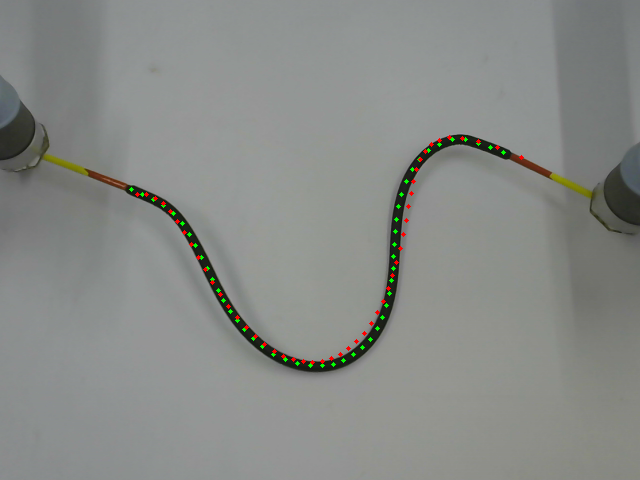}}

\subfloat[KMS]
{\includegraphics[scale=0.19]{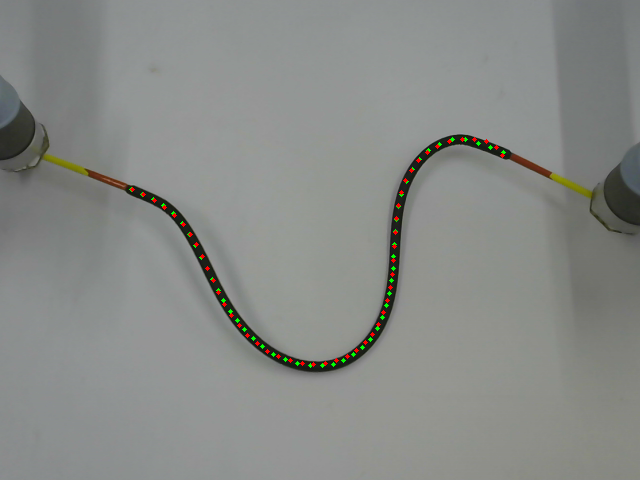}}
\subfloat[SOM]
{\includegraphics[scale=0.19]{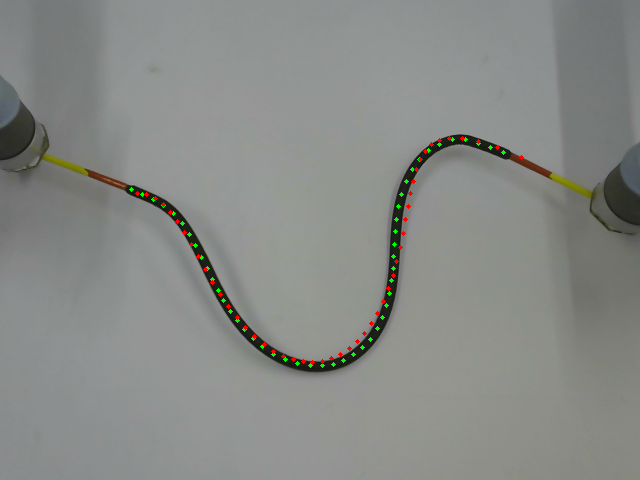}}
	
\caption{
Comparison of feature extraction effects within various fitting orders, i.e., $n=2, n=4,n=6$, and $n=8$.
}
\label{fig2}
\end{figure}

\begin{figure}[ht]
\centering
\includegraphics[scale=0.11]{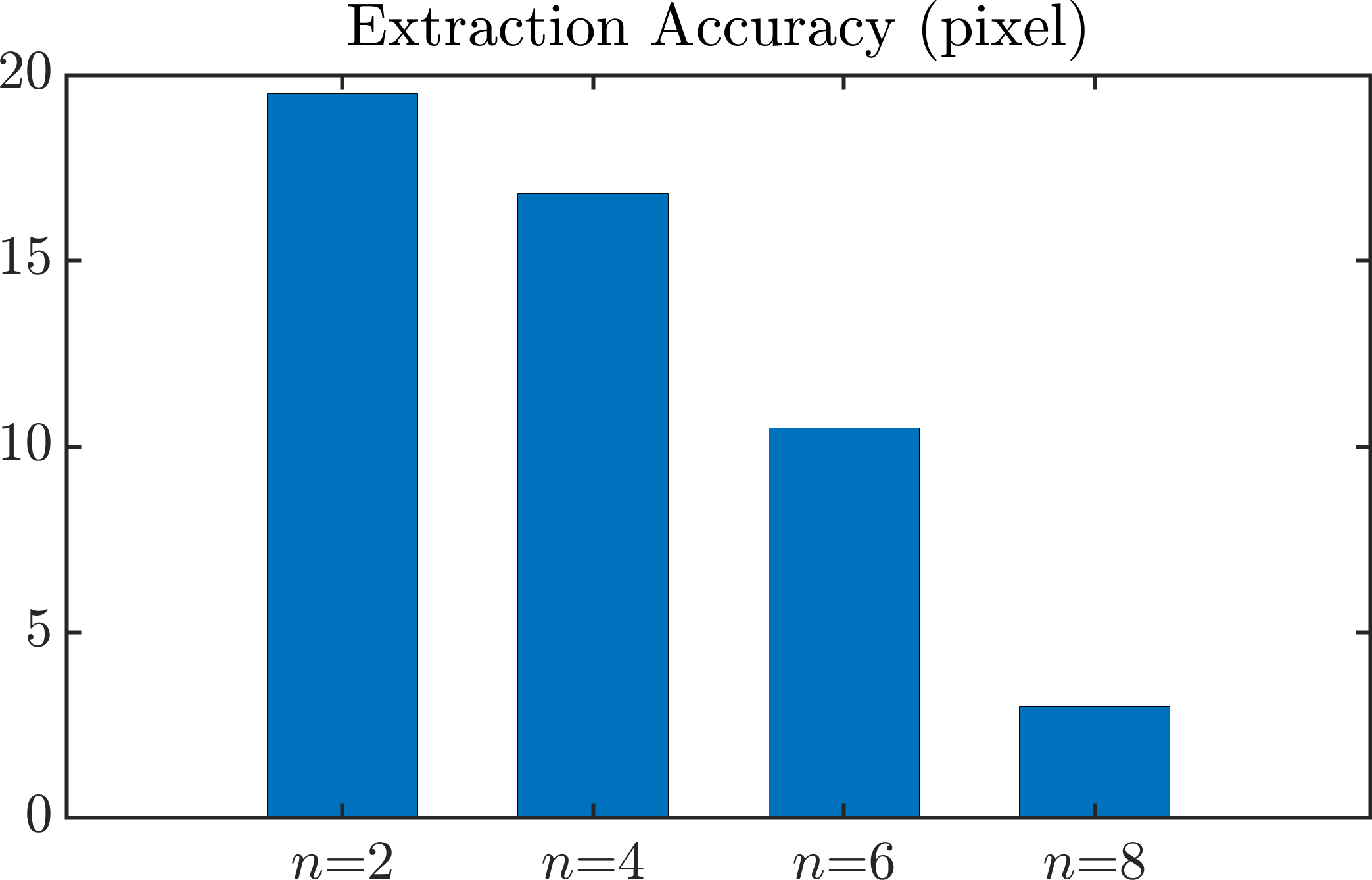}
\caption{
Fitting performance comparison of Bezier curve within various fitting orders.
}
\label{fig3}
\end{figure}

\subsection{Feature Extraction}
In this section, the proposed feature extraction module is tested within various fitting orders, i.e., $n=2, n=4, n=6, n=8$.
The number of shape points for the centerline configuration is set to $N=120$.
The reason why setting the fitting order $n$ is big is to ensure the validation of the assumption of the matrix invert ability.

Fig. \ref{fig3} presents the extraction comparison of the proposed Bezier-based feature extraction method.
The fitting order is set from $n=2$ to $n=8$.
The result shows that, as the fitting order $n$ increases, the extraction accuracy also increases.
And the case with $n=8$ is the best, followed by $n=6$ and $n=4$, while $n=2$ provides the worst performance.

\section{Conclusion}
In this paper, we presented a feature extraction approach to generate a low-dimensional feature that represents the whole shape configuration of the deformable linear object.
The proposed extraction algorithm successfully addresses the high-dimensional issue, which can help the robot to complete the manipulation and does not need any artificial markers.
Our experiments on real-world experiments, the results demonstrate that the approach introduced in this work has better accuracy and faster real-time performance than the existing methods, showcasing its effectiveness and robustness.

\appendices
\ifCLASSOPTIONcaptionsoff
  \newpage
\fi

\bibliography{biblio.bib}
\bibliographystyle{IEEEtran}
\end{document}